\pdfoutput=1

\documentclass[11pt,a4paper]{article}
\usepackage{times,latexsym}
\usepackage{url}
\usepackage[T1]{fontenc}
\usepackage[dvipsnames]{xcolor}
\usepackage{amssymb}
\usepackage{amsmath}
\usepackage{dsfont}
\usepackage{enumitem}
\usepackage{booktabs}
\usepackage{array}
\usepackage{tikz}
\usetikzlibrary{decorations.pathreplacing,calc,tikzmark,calligraphy,positioning,arrows.meta}
\usepackage{bm}
\usepackage{prftree}
\usepackage[ruled,linesnumbered]{algorithm2e}
\usepackage[defaultsans]{lato}
\usepackage{float}
\usepackage{colortbl}
\usepackage{titlesec}
\usepackage{footnotehyper}

\makeatletter
\newcommand{\removelatexerror}{\let\@latex@error\@gobble}

\g@addto@macro\normalsize{%
  \setlength\abovedisplayskip{3pt}
  \setlength\belowdisplayskip{3pt}
  \setlength\abovedisplayshortskip{3pt}
  \setlength\belowdisplayshortskip{3pt}
}
\makeatother

\colorlet{mdtRed}{red!70!black}
\definecolor{pale_red}{rgb}{0.94,0.92,0.92}

\titlespacing*{\section}{0pt}{0.5\baselineskip}{0.25\baselineskip}
\titlespacing*{\subsection}{0pt}{0.4\baselineskip}{0.25\baselineskip}
\titlespacing*{\subsubsection}{0pt}{0.3\baselineskip}{0.2\baselineskip}
\titleformat{\section}{\normalfont\fontsize{12}{15}\bfseries}{\thesection}{1em}{}
\titleformat{\subsection}{\normalfont\fontsize{11}{13}\bfseries}{\thesubsection}{1em}{}

\usepackage[acceptedWithA]{tacl2018v2}
\usepackage{breakurl}
\usepackage{cleveref}
\crefformat{footnote}{#2\footnotemark[#1]#3}

\title{Towards General Natural Language Understanding \\ with Probabilistic Worldbuilding}

\author{
 Abulhair Saparov \and Tom M. Mitchell \\
 Machine Learning Department \\
 Carnegie Mellon University \\
 {\sf \{asaparov, tom.mitchell\}@cs.cmu.edu}
}

\date{}

\begin{document}
\maketitle
\begin{abstract}
	\vspace{-2mm}
	We introduce the Probabilistic Worldbuilding Model (\textsf{PWM}), a new fully-symbolic Bayesian model of semantic parsing and reasoning, as a first step in a research program toward more domain- and task-general NLU and AI. Humans create internal mental models of their observations which greatly aid in their ability to understand and reason about a large variety of problems. In \textsf{PWM}, the meanings of sentences, acquired facts about the world, and intermediate steps in reasoning are all expressed in a human-readable formal language, with the design goal of interpretability. \textsf{PWM} is Bayesian, designed specifically to be able to generalize to new domains and new tasks. We derive and implement an inference algorithm that reads sentences by parsing and abducing updates to its latent world model that capture the semantics of those sentences, and evaluate it on two out-of-domain question-answering datasets: (1) \textsf{ProofWriter} and (2) a new dataset we call \textsf{FictionalGeoQA}, designed to be more representative of real language but still simple enough to focus on evaluating reasoning ability, while being robust against heuristics. Our method outperforms baselines on both, thereby demonstrating its value as a proof-of-concept.
\end{abstract}

\vspace{-1em}
\section{Introduction}
\vspace{-0.3em}

Despite recent progress in AI and NLP producing algorithms that perform well on a number of NLP tasks, it is still unclear how to move forward and develop algorithms that understand language as well as humans do. In particular, large-scale language models such as BERT \cite{DBLP:conf/naacl/DevlinCLT19}, RoBERTa \cite{DBLP:journals/corr/abs-1907-11692}, GPT-3 \cite{DBLP:journals/corr/abs-2005-14165}, XLNet \cite{DBLP:conf/nips/YangDYCSL19}, and others were trained on a very large amount of text and can then be applied to perform many different NLP tasks after some fine-tuning. In the case of GPT-3, some tasks require very few additional training examples to achieve state-of-the-art performance. As a result of training on text from virtually every domain, these models are domain-general. This is in contrast with NLP algorithms that are largely trained on one or a small handful of domains, and as such, are not able to perform well on new domains outside of their training. Despite this focus on domain-generality, there are still a large number of tasks on which these large-scale language models perform poorly \cite{DBLP:conf/acl/DunietzBBRCF20}. Many limitations of today's state-of-the-art methods become evident when comparing with the human ability to understand language \cite{DBLP:journals/corr/LakeUTG16,DBLP:conf/acl/TamariSHPAS20,DBLP:conf/acl/BenderK20,DBLP:conf/acl-mrqa/GardnerBHTM19,DBLP:conf/acl/Linzen20}. Many cognitive scientists posit that humans create rich mental models of the world from their observations which provide superior explainability, reasoning, and generalizability to new domains and tasks. How do we, as a field, move from today's state-of-the-art to more general intelligence? What are the next steps to develop algorithms that can generalize to new tasks at the same level as humans? The lack of interpretability in many of these models makes these questions impossible to answer precisely. One promising direction is to change the evaluation metric: \citet{DBLP:journals/corr/abs-2005-14165}, \citet{DBLP:conf/acl/Linzen20}, and many others have suggested \emph{zero-shot} or \emph{few-shot accuracy} to measure the performance of algorithms (i.e. the algorithm is evaluated with a new dataset, wholly separate from its training; or in the case of few-shot learning, save for a few examples). While this shift is welcome, it alone will not solve the above issues.

\begin{figure*}[h]
\centering
\footnotesize
\vspace{-2mm}
\begin{tikzpicture}[remember picture]
	\linespread{0.8}
	\fontfamily{ppl}\selectfont
	\node[fill=pale_red,align=center,rounded corners=2pt,label={[label distance=-0.2em]\textbf{\textsf{\small Theory}}}] (theory) {\hspace{0.4em}\begin{minipage}[c]{2em}$T \hspace{0.3em}= $\end{minipage}
		\begin{minipage}{32.5em}$\begin{array}{l}
			\{ \hspace{0.9em} {\color{RoyalBlue}\texttt{mammal}(\texttt{alice})}, \hspace{1.7em} {\color{RoyalBlue}\texttt{cat}(\texttt{bob})}, \hspace{1.7em} {\color{RoyalBlue}\forall x(\texttt{cat}(x) \to \texttt{mammal}(x))}, \hspace{0.9em} \hdots \hspace{0.9em} \}
		\end{array}$\end{minipage}
	};
	\node[fill=pale_red,align=center,rounded corners=2pt,below=2.4em of theory,label={[label distance=-0.2em]\textbf{\textsf{\small Proof and logical form of $\bm{i^{\text{\tiny th}}}$ sentence}}}] (proof) {\hspace{0.4em}\begin{minipage}[c]{2em}$\pi_i = $\end{minipage}
		\begin{minipage}{22.1em}\hspace{0.5em}
			\prftree[r]{$\to\hspace{-0.2em}\text{E}$}{
				\prftree[r]{Ax}{\color{RoyalBlue} \texttt{cat}(\texttt{bob})}
			}{
				\prftree[r]{$\forall\text{E}$}{
					\prftree[r]{Ax}{\color{RoyalBlue} \forall x(\texttt{cat}(x) \to \texttt{mammal}(x))}
				}{\color{RoyalBlue} \texttt{cat}(\texttt{bob}) \to \texttt{mammal}(\texttt{bob})}
			}{\color{RoyalBlue} \texttt{mammal}(\texttt{bob})}
		\end{minipage}
	};
	\node[fill=pale_red,align=center,rounded corners=2pt,below=2.1em of proof,label={[label distance=-0.1em]\textbf{\textsf{\small $\bm{i^{\text{\tiny th}}}$ sentence}}}] (sentence) {\hspace{0.3em}\begin{minipage}[c]{2em}$y_i = $\end{minipage}
		\begin{minipage}[c]{9.8em}\hspace{0.2em} ``Bob is a mammal.''\end{minipage}};

	\draw [-{Latex[length=1.8mm,width=1.8mm]},thick,rounded corners=2pt,line width=1.3pt,mdtRed!73] ([xshift=-16.5em,yshift=-0.3em]theory.south) |- ([xshift=-0.3em]proof.west) node [anchor=east,text width=6.8em,xshift=-3.4em,yshift=1em,align=right] {\scriptsize\color{black!65} generate proofs from the axioms in the theory};
	\draw [-{Latex[length=1.8mm,width=1.8mm]},thick,rounded corners=2pt,line width=1.3pt,mdtRed!73] ([xshift=-11em,yshift=-0.3em]proof.south) |- ([xshift=-0.3em]sentence.west) node [anchor=east,text width=9em,xshift=-4.5em,yshift=0.7em,align=right] {\scriptsize\color{black!65} generate sentence from logical form};

	\draw [{Latex[length=1.8mm,width=1.8mm]}-,thick,rounded corners=2pt,line width=1.3pt,OliveGreen!73] ([xshift=11em,yshift=-0.3em]proof.south) |- ([xshift=0.3em]sentence.east) node [anchor=west,text width=6em,xshift=4.5em,yshift=0.7em] {\scriptsize\color{black!65} parse sentence to logical form};
	\draw [{Latex[length=1.8mm,width=1.8mm]}-,thick,rounded corners=2pt,line width=1.3pt,OliveGreen!73] ([xshift=16.5em,yshift=-0.3em]theory.south) |- ([xshift=0.3em]proof.east) node [anchor=west,text width=9em,xshift=3.4em,yshift=1em] {\scriptsize\color{black!65} infer theory and proofs from observed logical forms};
\end{tikzpicture}
\vspace{-5mm}
\caption{The {\color{mdtRed}generative process} and {\color{OliveGreen}inference} in our model, with an example of a theory, generating a proof of a logical form which itself generates the sentence ``Bob is a mammal.'' During inference, only the sentences are observed, whereas the theory and proofs are latent. Given sentence $y_i$, the language module outputs the logical form. The reasoning module then infers the proof $\pi_i$ of the logical form and updates the posterior of the theory $T$.} \label{fig:generative_process}
\vspace{-5mm}
\end{figure*}
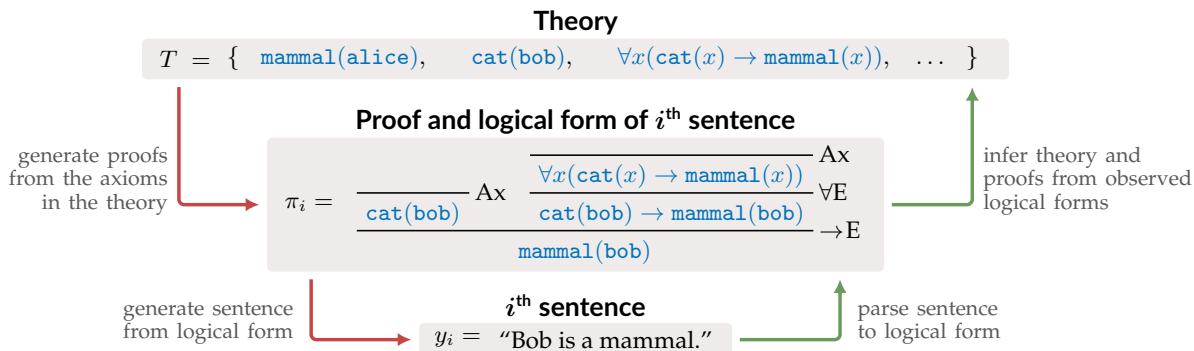

We introduce the \emph{Probabilistic Worldbuilding Model} (\textsf{PWM}), a probabilistic generative model of reasoning and semantic parsing. Like some past approaches, \textsf{PWM} explicitly builds an internal mental model, which we call the \emph{theory} \cite{DBLP:conf/acl/TamariSHPAS20,DBLP:journals/csur/HoganBCdMGKGNNN21,DBLP:journals/cacm/MitchellCHTYBCM18,DBLP:journals/ai/CharniakG93}. The theory constitutes what the algorithm believes to be true. \textsf{PWM} is fully symbolic and Bayesian, using a single unified human-readable formal language to represent all meaning, and is therefore \emph{inherently interpretable}. This is in contrast to systems that use subsymbolic representations of meaning for some or all of their components. Every random variable in \textsf{PWM} is well-defined with respect to other random variables and/or grounded primitives. Prior knowledge such as the rules of deductive inference, the structure of English grammar, knowledge of basic physics and mathematics can be incorporated by modifying the prior distributions of the random variables in \textsf{PWM}. Incorporating prior knowledge can greatly reduce the amount of training data required to achieve sufficient generalizability, as we will demonstrate. Extensibility is key to future research that could enable more general NLU and AI, as it provides a clearer path forward for future exploration.

We present an implementation of inference under the proposed model, called the \emph{Probabilistic Worldbuilding from Language} (\textsf{PWL}). While \textsf{PWM} is an abstract mathematical description of the underlying distribution of axioms, proofs, logical forms, and sentences, \textsf{PWL} is the algorithm that reads sentences, computes logical form representations of their meaning, and updates the axioms and proofs in the theory accordingly. See figure \ref{fig:generative_process} for a high-level schematic diagram of \textsf{PWM} and \textsf{PWL}. \textsf{PWM} describes the process depicted by the {\color{mdtRed}red arrows}, whereas \textsf{PWL} is the algorithm depicted by the {\color{OliveGreen}green arrows}. We emphasize that the reasoning in \textsf{PWL} is not a theorem prover and is not purely deductive. Instead, \textsf{PWL} solves a different problem of \emph{abduction}, which in some ways is computationally easier than deductive inference: Given a set of observations, work backwards to find a set of axioms that deductively \emph{explain} the observations. It is these abduced axioms that constitute the internal ``mental model.'' Humans often rely on abductive reasoning, for example in commonsense reasoning \cite{DBLP:conf/iclr/BhagavatulaBMSH20,DBLP:conf/cade/FurbachGS15}.

A core principle of our approach is to ensure \emph{generality by design}. Simplifying assumptions often trade away generality for tractability, such as by restricting the representation of the meanings of sentences, or number of steps during reasoning. \textsf{PWM} is designed to be domain- and task-general, and to this end, uses higher-order logic (i.e. lambda calculus) \cite{DBLP:journals/jsyml/Church40} as the formal language, which we believe is sufficiently expressive to capture the meaning of declarative and interrogative sentences in natural language. Furthermore, \textsf{PWM} uses \emph{natural deduction} for reasoning, which is \emph{complete} in that if a logical form $\phi$ is true, there is a proof of $\phi$ \cite{DBLP:journals/jsyml/Henkin50}.

In section \ref{sec:model}, we describe \textsf{PWM} and \textsf{PWL} more precisely. In section \ref{sec:experiments}, as a proof-of-concept of the value of further research, we run experiments on two question-answering datasets: \textsf{ProofWriter} \cite{DBLP:conf/acl/TafjordDC21} and a new dataset, called \textsf{FictionalGeoQA}, which we specifically created to evaluate the ability to reason over short paragraphs while being robust against simpler heuristic strategies. Unlike \textsf{ProofWriter}, the text in \textsf{FictionalGeoQA} was not template-generated and is more realistic, but is still simple enough to focus the evaluation on reasoning rather than parsing, with many sentences having semantics that go beyond the Horn clause fragment of first-order logic. \textsf{PWL} outperforms the baselines with respect to zero-shot accuracy (i.e. without looking at any training examples). Our code and data is freely available at \href{https://github.com/asaparov/PWL}{\texttt{\footnotesize github.com/asaparov/PWL}} and \href{https://github.com/asaparov/fictionalgeoqa}{\texttt{\footnotesize github.com/asaparov/fictionalgeoqa}}.

In summary, the primary contributions of this paper are the following:
\begin{itemize}[noitemsep,topsep=0.1pt,leftmargin=5mm]
	\item \textsf{PWM}, a new model for more general NLU, and \textsf{PWL}, the implementation that reads sentences, computes their logical forms, and updates its theory accordingly.
	\item Introducing \textsf{FictionalGeoQA}, a new question-answering dataset designed to evaluate the ability to reason over language.
	\item Experiments on \textsf{ProofWriter} and \textsf{FictionalGeoQA} demonstrating that \textsf{PWL} outperforms baseline methods on question-answering.
\end{itemize}

\section{Related work}

\emph{Fully symbolic} methods were commonplace in earlier AI research \cite{DBLP:journals/cacm/NewellS76,Dreyfus85}. However, they were oftentimes brittle: A new observation would contradict the internal theory or violate an assumption, and it was not clear how to resolve the impasse in a principled manner and proceed. But they do have some key advantages: Symbolic approaches that use well-studied human-readable formal languages such as first-order logic, higher-order logic, type theory, etc. enable humans to readily inspect and understand the internal processing of these algorithms, effecting a high degree of interpretability \cite{alma991019721785904436,10.3366/j.ctt1g09w89,cooper-etal-2015-probabilistic}. Symbolic systems can be made general by design, by using a sufficiently expressive formal language and ontology. Hybrid methods have been explored to alleviate the brittleness of formal systems while engendering their strengths, such as interpretability and generalizability; for example, the recent work in \emph{neuro-symbolic} methods \cite{DBLP:conf/iclr/YiGLK0TT20,DBLP:conf/emnlp/SahaGSB20,DBLP:conf/acl/TafjordDC21}. Neural theorem provers are in this vein \cite{DBLP:conf/nips/Rocktaschel017}. However, the proofs considered in these approaches are based on \emph{backward chaining} \cite{DBLP:books/daglib/0023820}, which restricts the semantics to the Horn clause fragment of first-order logic. \citet{DBLP:conf/nips/SunAB0C20,DBLP:conf/iclr/RenHL20,arakelyan2021complex} extend coverage to the existential positive fragment of first-order logic. In natural language, sentences express more complex semantics such as negation, nested universal quantification, and higher-order structures. Our work explores the other side of the tradeoff between tractability and expressivity/generality. Theorem provers attempt to solve the problem of deduction: finding a proof of a given formula, given a set of axioms. In contrast, the reasoning component of \textsf{PWM} is abductive, and the abduced axioms can be used in downstream tasks, such as question-answering, and to better read new sentences in the context of the world model, as we will demonstrate. We posit that abduction is sufficient for more general NLU \cite{Hobbs2008,DBLP:journals/ai/HobbsSAM93}. \textsf{PWM} combines Bayesian statistical machine learning with symbolic representations in order to handle uncertainty in a principled manner, ``smoothing out'' or ``softening'' the rigidity of a purely symbolic approach. In \textsf{PWM}, the internal theory is a random variable, so if a new observation is inconsistent with the theory, there may be other theories in the probability space that are consistent with the observation. The probabilistic approach provides a principled way to resolve these impasses.

\textsf{PWM} is certainly not the first to combine symbolic and probabilistic methods. There is a rich history of \emph{inductive logic programming} (ILP) \cite{DBLP:journals/ngc/Muggleton91,DBLP:journals/ml/CropperM21} and probabilistic ILP languages \cite{Muggleton1996,DBLP:journals/ml/Cussens01,DBLP:conf/ijcai/SatoKZ05,DBLP:journals/tplp/BellodiR15}. These languages could be used to learn a ``theory'' from a collection of observations, but they are typically restricted to learning rules in the form of first-order Horn clauses, for tractability. In natural language, it is easy to express semantics beyond the Horn clause fragment of first-order logic.

\emph{Knowledge bases} (KBs) and \emph{cognitive architectures} \cite{DBLP:journals/air/KotserubaT20,DBLP:journals/csur/HoganBCdMGKGNNN21,DBLP:journals/ai/LairdNR87,DBLP:journals/cacm/MitchellCHTYBCM18} have attempted to explicitly model domain-general knowledge in a form amenable to reasoning. Cognitive architectures aim to more closely replicate human cognition. Some approaches use probabilistic methods to handle uncertainty \cite{DBLP:conf/naacl/NiepertMS12,DBLP:conf/uai/NiepertD15,DBLP:conf/akbc/JainFKBR19}. However, many of these approaches make strong simplifying assumptions that restrict the expressive power of the formal language that expresses facts in the KB. For example, many knowledge bases can be characterized as graphs, where each entity corresponds to a vertex and every fact corresponds to a labeled edge. For example, the belief \texttt{\color{RoyalBlue}plays\_sport(s\_williams,tennis)} is representable as a directed edge connecting the vertex \texttt{\color{RoyalBlue}s\_williams} to the vertex \texttt{\color{RoyalBlue}tennis}, with the edge label \texttt{\color{RoyalBlue}plays\_sport}. While this assumption greatly aids tractability and scalability, allowing many problems in reasoning to be solved by graph algorithms, it greatly hinders expressivity and generality, and there are many kinds of knowledge that simply cannot be expressed and represented in such KBs. \textsf{PWM} does not make such restrictions on logical forms in the theory, allowing for richer semantics, such as definitions, universally-quantified statements, conditionals, etc.

\section{Model} \label{sec:model}

In this section, we provide a mathematical description of \textsf{PWM}. At a high level, the process for generating a sentence sampled from this probability distribution is:
\begin{enumerate}[noitemsep,topsep=0.1pt,leftmargin=5mm]
	\item Sample the theory $T$ from a prior distribution $p(T)$. $T$ is a collection of logical forms in higher-order logic that represent what \textsf{PWL} believes to be true.
	\item For each observation $i$, sample a proof $\pi_i$ from $p(\pi_i \mid T)$. The conclusion of the proof is the logical form $x_i$, which represents the meaning of the $i^{\text{\scriptsize th}}$ sentence.
	\item Sample the $i^{\text{\scriptsize th}}$ sentence $y_i$ from $p(y_i \mid x_i)$.
\end{enumerate}
Inference is effectively the inverse of this process, and is implemented by \textsf{PWL}. During inference, \textsf{PWL} is given a collection of observed sentences $y_1,\ldots,y_n$ and the goal is to discern the value of the latent variables: the logical form of each sentence $\bm{x}\triangleq\{x_1,\ldots,x_n\}$, the proofs for each logical form $\bm{\pi}\triangleq\{\pi_1,\ldots,\pi_n\}$, and the underlying theory $T$. Both the generative process and inference algorithm naturally divide into two modules:
\begin{itemize}[noitemsep,topsep=0.1pt,leftmargin=5mm]
	\item \textbf{Language module:} During inference, this module's purpose is to infer the logical form of each observed sentence. That is, given the input sentence $y_i$, this module outputs the $k$ most-probable values of the logical form $x_i$ (i.e. semantic parsing).
	\item \textbf{Reasoning module:} During inference, this module's purpose is to infer the underlying theory that logically entails the observed logical forms (and their proofs thereof). That is, given an input collection of logical forms $\bm{x}$, this module outputs the posterior distribution of the underlying theory $T$ and the proofs $\bm{\pi}$ of those logical forms.
\end{itemize}
Note that the $y_i$ need not necessarily be sentences, and \textsf{PWM} can easily be generalized to other kinds of data. For example, if a generative model of images is available for $p(y_i \mid x_i)$, then an equivalent "vision module" may be defined. This module may be used either in place of, or together with the language module. In the above generative process, \textsf{PWM} assumes each sentence to be independent. A model of context is required to properly handle inter-sentential anaphora or conversational settings. This can be done by allowing the distribution on $y_i$ to depend on previous logical forms or sentences: $p(y_i \mid x_1,\ldots,x_i)$ (i.e. relaxing the i.i.d. assumption). For simplicity of this proof-of-concept, this is left to future work.

\begin{table*}[h!]
	\centering
	\renewcommand{\arraystretch}{1.2}
	\newcommand{\Land}{\hspace{0.2em}\land\hspace{0.2em}}
	\scriptsize
	\setlength\aboverulesep{0.605mm}
	\setlength\belowrulesep{0.605mm}
	\hspace{1em}\begin{tabular}{ >{\raggedright}m{0.24\textwidth} >{\raggedright}m{0.36\textwidth} >{\raggedright}m{0.22\textwidth} }
		\toprule
		& \hspace{3em}\textbf{Semantic parser output} & \textbf{Possible axioms in theory} \tabularnewline \toprule
			Without neo-Davidsonian semantics, language module performs entity linking &
			$\color{RoyalBlue}\exists b (\texttt{book}(b) \Land \texttt{write}(\texttt{alex},b))$ &
			$\color{RoyalBlue}\texttt{book}(c_1)$, $\color{RoyalBlue}\texttt{write}(\texttt{alex},c_1)$. \tabularnewline \midrule
			\emph{With} neo-Davidsonian semantics, language module performs entity linking &
			$\color{RoyalBlue}\exists b (\texttt{book}(b) \Land \exists w(\texttt{write}(w)$
				$\color{RoyalBlue} \hspace{0.5em}\Land \texttt{arg1}(w)=\texttt{alex} \Land \texttt{arg2}(w)=b))$ &
			$\color{RoyalBlue}\texttt{book}(c_1)$, $\color{RoyalBlue}\texttt{write}(c_2)$, $\color{RoyalBlue}\texttt{arg1}(c_2)=\texttt{alex}$, $\color{RoyalBlue}\texttt{arg2}(c_2)=c_1$. \tabularnewline \midrule
			\tikzmark{A} \hspace{-0.6em} \emph{With} neo-Davidsonian semantics, reasoning module resolves named entities \newline \vspace{-1.5em} \tikzmark{B} &
			$\color{RoyalBlue}\exists a (\texttt{name}(a)=\textrm{``Alex''}$
				$\color{RoyalBlue}\hspace{0.5em}\Land \exists b (\texttt{book}(b) \Land \exists w(\texttt{write}(w)$
				$\color{RoyalBlue}\hspace{1.3em}\Land \texttt{arg1}(w)=a \Land \texttt{arg2}(w)=b)))$ &
			$\color{RoyalBlue}\texttt{book}(c_1)$, $\color{RoyalBlue}\texttt{write}(c_2)$, $\color{RoyalBlue}\texttt{name}(c_3)=\textrm{``Alex''}$, $\color{RoyalBlue}\texttt{arg1}(c_2)=c_3$, $\color{RoyalBlue}\texttt{arg2}(c_2)=c_1$. \tabularnewline \bottomrule
	\end{tabular}
	\begin{tikzpicture}[remember picture,overlay]
		\draw [thick,decorate,decoration={calligraphic brace,amplitude=10pt,raise=4pt}]
		($(pic cs:B) + (-0.2,-0.02)$) -- ($(pic cs:A) + (-0.13, 0.48)$) node [midway,xshift=-0.8cm,rotate=90] {our approach};
	\end{tikzpicture}
	\vspace{-2mm}
	\caption{Design choices in the representation of the meaning of ``Alex wrote a book.''} \label{fig:semantic_representations}
	\vspace{-2em}
\end{table*}

There is a vast design space for symbolic representations of meaning. We are unable to comprehensively list all of our design choices, but we describe two important ones below.

Neo-Davidsonian semantics \cite{Parsons1990} is used to represent meaning in all logical forms (both in the theory and during semantic parsing). As a concrete example, a straightforward way to represent the meaning of ``Jason traveled to New York'' could be with the logical form ${\color{RoyalBlue}\texttt{travel}(\texttt{jason},\texttt{nyc})}$. In neo-Davidsonian semantics, this would instead be represented with three distinct atoms: ${\color{RoyalBlue}\texttt{travel}(c_1)}$, ${\color{RoyalBlue}\texttt{arg1}(c_1)=\texttt{jason}}$, and ${\color{RoyalBlue}\texttt{arg2}(c_1)=\texttt{nyc}}$. Here, $c_1$ is a constant that represents the ``traveling event,'' whose first argument is the constant representing Jason, and whose second argument is the constant representing New York City. This representation allows the event to be more readily modified by other logical expressions, such as in ``Jason quickly traveled to NYC before nightfall.''

In addition, \textsf{PWM} defers named entity linking to the reasoning module (it is not done during parsing). That is, the semantic parser does not parse ``Jason'' directly into the constant $\color{RoyalBlue}\texttt{jason}$. Rather, named entities are parsed into existentially-quantified expressions, e.g. $\color{RoyalBlue}\exists j (\texttt{name}(j)=\textrm{``Jason''} \land \ldots)$. This simplifies the parser's task and allows reasoning to aid in entity linking. Table \ref{fig:semantic_representations} details these design options.

\subsection{Reasoning module}

\subsubsection{Generative process for the theory $\bm{p(T)}$}

The theory $T$ is a collection of axioms $a_1, a_2, \ldots$ represented in higher-order logic. We choose a fairly simple prior $p(T)$ for rapid prototyping, but it is straightforward to substitute with a more complex prior. Specifically $a_1, a_2, \ldots$ are distributed according to a distribution $G_a$ which is sampled from a \emph{Dirichlet process} (DP) \cite{Ferguson73}, an exchangeable non-parametric distribution.
\begin{align}
	G_a &\sim \text{DP}(H_a, \alpha), \\
	a_1,a_2,\ldots &\sim G_a,
\end{align}
where $H_a$ is the \emph{base distribution} and $\alpha = 0.1$ is the concentration parameter. An equivalent perspective of the DP that better illustrates how the samples are generated is the \emph{Chinese restaurant process} \cite{Aldous1985}:
\begin{align}
	\hspace{-0.3em} a_i &\sim H_a \text{ with probability } \frac{\alpha}{\alpha + i - 1}, \\
	\hspace{-0.3em} a_i &= a_j \text{ with probability } \frac{\sum_{k=1}^{i-1} \mathds{1}\{a_k = a_j\}}{\alpha + i - 1}.
\end{align}
That is, the $i^{\text{\scriptsize th}}$ sample is drawn from $H_a$ with probability proportional to $\alpha$, or it is set to a previous sample with probability proportional to the number of times that sample has been drawn.

The base distribution $H_a$ recursively generates logical forms in higher-order logic. Since any formula can be written as a tree, they can be generated top-down, starting from the root. The type of each node (i.e. conjunction $\color{RoyalBlue}\land$, disjunction $\color{RoyalBlue}\lor$, negation $\color{RoyalBlue}\neg$, quantification $\color{RoyalBlue}\forall x$, etc) is sampled from a categorical distribution. If the type of the node is selected to be an atom (e.g. $\color{RoyalBlue}\texttt{book}(c_1)$), then its predicate is sampled from a non-parametric distribution of predicate symbols $H_p$. The atom's argument(s) are each sampled as follows: if $n_V$ is the number of available variables (from earlier generated quantifiers), then sample a variable uniformly at random with probability $\smash{\frac{1}{n_V+1}}$; otherwise, with probability $\smash{\frac{1}{n_V+1}}$, sample a constant from a non-parametric distribution of constant symbols $H_c$. For brevity, we refer the reader to our code for the specific forms of $H_p$ and $H_c$.

Since \textsf{PWM} uses a neo-Davidsonian representation, another node type that $H_a$ can generate is an event argument (e.g. ${\color{RoyalBlue}\texttt{arg1}(c_1)=\texttt{jason}}$). When this is selected, the event constant ($\color{RoyalBlue}c_1$ in the example) is sampled in the same way an atom's argument is sampled, as described above: first by trying to sample a variable, and otherwise sampling a constant from $H_c$. The right-side of the equality ($\color{RoyalBlue}\texttt{jason}$ in the example) can either be a variable, constant, string, or number, so \textsf{PWM} first selects its type from a categorical distribution. If the type is chosen to be a number, string, or variable, its value is sampled uniformly. If the type is chosen to be a constant, it is sampled from $H_c$.

\emph{Names of entities} are treated specially in this prior: The number of names available to each entity is sampled according to a very light-tailed distribution i.i.d.: for entity $c_i$ the number of names $n_N(c_i) \triangleq \#\{s : {\color{RoyalBlue}\texttt{name}(c_i)=s}\}$ is distributed according to $p(n_N(c_i) = k) \propto \lambda^{k^2}$. This ensures that entities tend not to have too many names.

\emph{Sets} are also treated specially in this prior: One kind of axiom that can be generated is one that declares the size of a set, e.g. ${\color{RoyalBlue}\texttt{size}(\lambda x.\texttt{planet}(x))=8}$ denotes that the size of the set of planets is $8$. In the prior, the size of each set is distributed according to a geometric distribution with parameter $10^{-4}$. Sets can have arity not equal to 1, in which case their elements are tuples.

\phantomsection \label{section:deterministic_constraints}
\textbf{Deterministic constraints:} We also impose hard constraints on the theory $T$. Most importantly, $T$ is required to be \emph{globally consistent}. While this is a conceptually simple requirement, it is computationally expensive (generally undecideable even in first-order logic). \textsf{PWL} enforces this constraint by keeping track of the known sets in the theory (i.e. a set is known if its set size axiom is used in a proof, or if the set appears as a subset/superset in a universally-quantified axiom, such as in ${\color{RoyalBlue}\forall x(\texttt{cat}(x) \to \texttt{mammal}(x))}$ where the set ${\color{RoyalBlue}\lambda x.\texttt{cat}(x)}$ is a subset of ${\color{RoyalBlue}\lambda x.\texttt{mammal}(x)}$). For each set, \textsf{PWL} computes which elements are provably members of that set. If the number of provable members of a set is greater than its size, or if an element is both provably a member and not a member of a set, the theory is inconsistent. Relaxing this constraint would be valuable in future research, perhaps instead by only considering the \emph{relevant} sets rather than all sets in the theory, or deferring consistency checks altogether. We place a handful of other constraints on the theory $T$: The name of an entity must be a string (and not a number or a constant). All constants are distinct; that is, ${\color{RoyalBlue}c_i} \neq {\color{RoyalBlue}c_j}$ for all $i \neq j$. This helps to alleviate identifiability issues, as otherwise, there would be a much larger number of logically-equivalent theories. No event can be an argument of itself (e.g. there is no constant $\color{RoyalBlue}c_i$ such that ${\color{RoyalBlue}\texttt{arg1}(c_i)=c_i}$). If a theory $T$ satisfies all constraints, we write ``$T \text{ valid}$.''

These constraints do slightly complicate computation of the prior, since the generative process for generating $T$ is \emph{conditioned} on $T$ being valid:
\begin{align}
	p(T \mid T \text{ valid}) &= p(T) / p(T \text{ valid}), \label{eq:conditional_theory_prior} \\
	\text{where } p(T \text{ valid}) &= \sum_{T' : T' \text{ valid}} p(T'),
\end{align}
and the denominator is intractable to compute. However, we show in section \ref{section:reasoning_inference} that for inference, it suffices to be able to efficiently compute the ratio of prior probabilities:
\begin{equation}
	\hspace{-0.6em}\frac{p(T_1 | T_1 \text{ valid})}{p(T_2 | T_2 \text{ valid})} \hspace{-0.1em}=\hspace{-0.1em} \frac{p(T_1)p(T_2 \text{ valid})}{p(T_2)p(T_1 \text{ valid})} \hspace{-0.1em}=\hspace{-0.1em} \frac{p(T_1)}{p(T_2)}. \label{eq:conditional_theory_prior_ratio}
\end{equation}

\noindent Additionally note that since the above constraints do not depend on the \emph{order} of the axioms, constants, etc. (i.e. the constraints themselves are exchangeable), the distribution of $T$ conditioned on $T$ being valid is exchangeable.

\textbf{Properties of the prior $\bm{p(T)}$:} We emphasize that these distributions were chosen for simplicity and ease of implementation, and they worked well enough in experiments. However, there are likely many distributions that would work just as well. The parameters in the above distributions are not learned; they were set and fixed a priori. Nevertheless, this prior does exhibit useful properties for a domain- and task-general model of reasoning:
\begin{itemize}[noitemsep,topsep=0.1pt,leftmargin=5mm]
	\item \emph{Occam's razor:} Smaller/simpler theories are given higher probability than larger and more complex theories.
	\item \emph{Consistency:} Inconsistent theories are discouraged or impossible.
	\item Entities tend to have a unique name. Our prior above encodes one direction of this prior belief: each entity is unlikely to have many names. However, the prior does not discourage one name from referring to multiple entities.
	\item Entities tend to have a unique type. Note however that this does not discourage types provable by subsumption. For example, if the theory has the axioms ${\color{RoyalBlue}\texttt{novel}(c_1)}$ and $\color{RoyalBlue}\forall x(\texttt{novel}(x) \to \texttt{book}(x))$, even though ${\color{RoyalBlue}\texttt{book}(c_1)}$ is provable, it is not an axiom in this example and the prior only applies to axioms.
\end{itemize}

\subsubsection{Generative process for proofs $\bm{p(\pi_i \mid T)}$}

\textsf{PWM} uses \emph{natural deduction}, a well-studied proof calculus, for the proofs \cite{Gentzen1935,GEN69a}. \citet{Pfenning} provides an accessible introduction. Figure \ref{fig:natural_deduction_example} illustrates a simple example of a natural deduction proof. Each horizontal line is a deduction step, with the (zero or more) formulas above the line being its \emph{premises}, and the one formula below the line being its conclusion. Each deduction step has a label to the right of the line. For example, the ``$\land\text{I}$'' step denotes \emph{conjunction introduction}: given that $A$ and $B$ are true, this step concludes that $A\land B$ is true, where $A$ and $B$ can be any formula. A natural deduction proof can rely on axioms (denoted by ``Ax'').

\begin{figure}[h]
	\vspace{-1.9em}
	\centering
	\footnotesize
	\begin{equation*}
		\prftree[r]{$\neg\text{I}$}{
			\prftree[r]{$\neg\text{E}$}{
				\prftree[r]{$\land\text{E}$}{
					\prftree[r]{Ax}{A\land \neg A}
				}{A}
			}{
				\prftree[r]{$\land\text{E}$}{
					\prftree[r]{Ax}{A\land \neg A}
				}{\neg A}
			}{\bot}
		}{\neg(A\land \neg A)}
	\end{equation*}
	\vspace{-2.9em}
	\caption{An example of a proof of $\neg(A\land \neg A)$.}
	\label{fig:natural_deduction_example}
	\vspace{-1.2em}
\end{figure}

\noindent We can write any natural deduction proof $\pi_i$ as a sequence of deduction steps $\pi_i \triangleq (\pi_{i,1},\ldots,\pi_{i,k})$ by traversing the proof tree in prefix order. We define a simple generative process for $\pi_i$:
\begin{enumerate}[noitemsep,topsep=0pt,leftmargin=5mm]
	\item First sample the length of the proof $k$ from a Poisson distribution with parameter $20$.
	\item For each ${j=1,\ldots,k}$: Select a deduction rule from the proof calculus with a categorical distribution. If the $\text{Ax}$ rule is selected, then simply take the next available axiom from the theory $T = a_1,a_2,\ldots$ If the deduction rule requires premises, then each premise is selected uniformly at random from $\pi_{i,1},\hdots,\pi_{i,j-1}$.\footnote{Some deduction rules require additional parameters, and we refer the reader to our code for details on how these parameters are sampled.}
\end{enumerate}
The above generative process may produce a forest rather than a single proof tree. Thus, $\pi_i$ is sampled conditioned on $\pi_i$ being a valid proof. Just as with $p(T)$ in equation \ref{eq:conditional_theory_prior}, this conditioning causes $p(\pi_i \mid T)$ to be intractable to compute. However, only the ratio of the prior probability is needed for inference, which can be computed efficiently:
\begin{align}
	&\frac{p(\pi_i \mid T, \pi_i \text{ valid})}{p(\pi_i' \mid T, \pi_i' \text{ valid})} \nonumber \\
	&\hspace{0.2em} = \hspace{0.1em} \frac{p(\pi_i\mid T)p(\pi_i' \text{ valid}\mid T)}{p(\pi_i'\mid T)p(\pi_i \text{ valid}\mid T)} = \frac{p(\pi_i\mid T)}{p(\pi_i'\mid T)}. \label{eq:conditional_proof_prior_ratio}
\end{align}

While \textsf{PWL} was initially implemented assuming classical logic, it is easy to adapt \textsf{PWL} to use other logics, such as \emph{intuitionistic logic}. Intuitionistic logic is identical to classical logic except that the \emph{law of the excluded middle} $A\lor\neg A$ is not a theorem (see figure \ref{fig:proofwriter_example} for an example where the two logics disagree). The interpretable nature of the reasoning module makes it easy to adapt it to other kinds of logic or proof calculi. \textsf{PWL} supports both classical and intuitionistic logic.

\subsubsection{Inference} \label{section:reasoning_inference}

Having described the generative process for the theory $T$ and proofs $\bm{\pi}$, we now describe inference. Given logical forms $\bm{x}$, the goal is to compute the posterior distribution of $T$ and $\bm{\pi}$ such that the conclusion of the each proof $\pi_i$ is $x_i$. That is, \textsf{PWL} aims to recover the latent theory and proofs that explain/entail the given observed logical forms. To this end, \textsf{PWL} uses Metropolis-Hastings (MH) \cite{hastings70,DBLP:books/sp/RobertC04}. \textsf{PWL} performs inference in a streaming fashion, starting with the case $n\! =\! 1$ to obtain MH samples from $p(\pi_1,T|x_1)$. Then, for every new logical form $x_n$, \textsf{PWL} uses the last sample from $p(\pi_1,\ldots,\pi_{n-1},T|x_1,\ldots,x_{n-1})$ as a starting point and then obtains MH samples from $p(\pi_1,\ldots,\pi_n,T|x_1,\ldots,x_n)$. This warm-start initialization serves to dramatically reduce the number of iterations needed to mix the Markov chain. To obtain the MH samples, the proof of each new logical form $\smash{\pi_n^{(0)}}$ is initialized using algorithm \ref{alg:init_proof}, whereas the proofs of previous logical forms are kept from the last MH sample. The axioms in these proofs constitute the theory sample $T^{(0)}$. Then, for each iteration $t\! =\! 1,\ldots,N_\text{iter}$, MH proposes a mutation to one or more proofs in $\bm{\pi}^{(t)}$. The possible mutations are listed in table \ref{table:mh_proposals}. This may change axioms in $T^{(t)}$. Let $T'$, $\pi_i'$ be the newly proposed theory and proofs. Then, compute the acceptance probability:
\begin{equation}
	\hspace{-0.2em}\frac{p(T')}{p(T^{(t)})} \prod_{i=1}^n \frac{p(\pi_i'| T')}{p(\pi_i^{(t)}| T^{(t)})} \frac{g(T^{(t)},\bm{\pi}^{(t)}| T',\bm{\pi}')}{g(T',\bm{\pi}'| T^{(t)},\bm{\pi}^{(t)})},
\end{equation}
where $g(T',\bm{\pi}'| T^{(t)},\bm{\pi}^{(t)})$ is the probability of proposing the mutation from $T^{(t)},\bm{\pi}^{(t)}$ to $T',\bm{\pi}'$, and $g(T^{(t)},\bm{\pi}^{(t)}| T',\bm{\pi}')$ is the probability of the \emph{inverse} of this mutation. Since this quantity depends only on the \emph{ratio} of probabilities, it can be computed efficiently (see equations \ref{eq:conditional_theory_prior_ratio} and \ref{eq:conditional_proof_prior_ratio}). Once this quantity is computed, sample from a Bernoulli with this quantity as its parameter. If it succeeds, MH accepts the proposed theory and proofs as the next sample: $\smash{T^{(t+1)}} \hspace{-0.2em}=\hspace{-0.2em} T'$ and $\smash{\pi_i^{(t+1)}} \hspace{-0.2em}=\hspace{-0.2em} \pi_i'$. Otherwise, reject the proposal and keep the old sample: $\smash{T^{(t+1)}} \hspace{-0.1em}=\hspace{-0.1em} \smash{T^{(t)}}$ and $\smash{\pi_i^{(t+1)}} \hspace{-0.1em}=\hspace{-0.1em} \smash{\pi_i^{(t)}}$. If every possible theory and proof is reachable from the initial theo- %

\pagebreak

\begingroup
\removelatexerror
\begin{algorithm}[H]
\scriptsize
\setstretch{0.95}
\let\oldnl\nl
\newcommand{\nonl}{\renewcommand{\nl}{\let\nl\oldnl}}
\SetNlSty{}{\color{RedOrange}\sffamily}{}
\SetAlgoBlockMarkers{}{}
\SetKwProg{Fn}{function}{}{}
\SetKwIF{If}{ElseIf}{Else}{if}{ }{else if}{else }{}
\SetKw{Continue}{continue}
\SetKwFunction{FInitProof}{\small init\_proof}
\SetKwFunction{FInitDisproof}{\small init\_disproof}
\SetKwProg{Fn}{function}{}{}
\AlgoDisplayBlockMarkers\SetAlgoVlined
\SetAlCapNameFnt{\small}
\SetAlCapFnt{\small}
\SetNoFillComment
\DontPrintSemicolon
\SetInd{0.0em}{0.8em}
	\Fn{\FInitProof{formula $A$}}{
		\uIf{$A$ is a conjunction $B_1 \land \ldots \land B_n$}{
			\For{$i=1$ to $n$}{$\phi_i = $ \texttt{init\_proof($B_i$)}}
			\Return{\hspace{0.2em}$\vcenter{
				\prftree[r]{$\land\text{I}$}{\phi_1}{\ldots}{\phi_n}{B_1 \land \ldots \land B_n}
			}$}
		}
		\uElseIf{$A$ is a disjunction $B_1 \lor \ldots \lor B_n$}{
			$\mathcal{I} = \texttt{shuffle($1,\ldots,n$)}$ \label{alg:init_proof:disjunction_intro} \;
			\For{$i\in\mathcal{I}$}{
				$\phi_i = $ \texttt{init\_proof($B_i$)} \;
				\If{$\phi_i \neq \text{null}$}{
					\Return{\hspace{0.2em}$\vcenter{
						\prftree[r]{$\lor\text{I}$}{\phi_i}{B_1 \lor \ldots \lor B_n}
					}$}
				}
			}
		}
		\uElseIf{$A$ is a negation $\neg B$}{
 			\Return{\normalfont\texttt{init\_disproof($B$)}}
		}
		\uElseIf{$A$ is an implication $B_1 \to B_2$}{
			\uIf{using classical logic}{
				$\mathcal{I} = \texttt{shuffle($1,2$)}$ \label{alg:init_proof:implication_intro} \;
				\For{$i\in\mathcal{I}$}{
					\uIf{$i = 1$}{
						$\phi_1 = $ \texttt{init\_disproof($B_1$)} \;
						\lIf{$\phi_1 = \text{null}$}{\Continue}
						\Return{\hspace{0.2em}$\vcenter{
							\prftree[r]{$\to\hspace{-0.2em}\text{I}$}{
								\prftree[r]{$\bot\text{E}$}{
									\prftree[r]{$\neg\text{E}$}{
										\phi_1
									}{
										\prftree[r]{Ax}{B_1}
									}{\bot}
								}{B_2}
							}{B_1 \to B_2}
						}$}
					}\Else{
						$\phi_2 = $ \texttt{init\_proof($B_2$)} \;
						\lIf{$\phi_2 = \text{null}$}{\Continue}
						\Return{\hspace{0.2em}$\vcenter{
							\prftree[r]{$\to\hspace{-0.2em}\text{I}$}{\phi_2}{B_1 \to B_2}
						}$}
					}
				}
			}\ElseIf{using intuitionistic logic}{
				\Return{\hspace{0.2em}$\vcenter{
					\prftree[r]{Ax}{B_1 \to B_2}
				}$}
			}
		}
		\uElseIf{$A$ is an existential quantification $\exists x.f(x)$}{
			let $\mathcal{C}$ be the set of known constants, numbers, and strings in $T$, and the new constant $c^*$ \;
			$\mathcal{I} = \texttt{swap\footnotemark(shuffle($\mathcal{C}$))}$ \label{alg:init_proof:existential_intro} \;
			\For{$c\in\mathcal{I}$}{ \label{alg:init_proof:existential_intro_loop}
				$\phi_c = $ \texttt{init\_proof($f(c)$)} \;
				\If{$\phi_c \neq \text{null}$}{
					\Return{\hspace{0.2em}$\vcenter{
						\prftree[r]{$\exists\text{I}$}{\phi_c}{\exists x.f(x)}
					}$}
				}
			}
		}
		\uElseIf{$A$ is a universal quantification $\forall x.f(x)$}{
			\Return{\hspace{0.2em}$\vcenter{
				\prftree[r]{Ax}{\forall x.f(x)}
			}$}
		}
		\uElseIf{$A$ is an equality $B_1 = B_2$}{
			\Return{\hspace{0.2em}$\vcenter{
				\prftree[r]{Ax}{B_1 = B_2}
			}$}
		}
		\ElseIf{$A$ is an atom (e.g. $\color{RoyalBlue}\texttt{book(great\_gatsby)}$)}{
			\Return{\hspace{0.2em}$\vcenter{
				\prftree[r]{Ax}{A}
			}$}
		}
		\lElse{\Return{null}}
	}
	\caption{Pseudocode for proof initialization. If any new axiom violates the deterministic constraints in section \ref{section:deterministic_constraints}, the function returns null.}
	\label{alg:init_proof}
\end{algorithm}
\endgroup

\dimen\footins=8\baselineskip\relax

\begin{savenotes}
\footnotetext{\texttt{swap} randomly selects an element in its input list to swap with the first element. The probability of moving an element $\color{RoyalBlue}c$ to the front of the list is computed as follows: Recursively inspect the atoms in the formula $\color{RoyalBlue}f(c)$ and count the number of ``matching'' atoms: The atoms $\color{RoyalBlue}t(c)$ or $\color{RoyalBlue}c(t)$ is considered ``matching'' if it is provable in $T$. Next, count the number of ``mismatching'' axioms: for each atom $\color{RoyalBlue}t(c)$ in the formula $\color{RoyalBlue}f(c)$, an axiom $\color{RoyalBlue}t'(c)$ is ``mismatching'' if ${\color{RoyalBlue}t}\neq {\color{RoyalBlue}t'}$. And similarly for each atom $\color{RoyalBlue}c(t)$ in the formula $\color{RoyalBlue}f(c)$, an axiom $\color{RoyalBlue}c(t')$ is ``mismatching'' if ${\color{RoyalBlue}t}\neq {\color{RoyalBlue}t'}$. Let $n$ be the number of ``matching'' atoms and $m$ be the number of ``mismatching'' axioms, then the probability of moving $c$ to the front of the list is proportional to $\exp\{n - 2m\}$. This greatly increases the chance of finding a high-probability proof in the first iteration of the loop on line \ref{alg:init_proof:existential_intro_loop}, and since this function is also used in an MH proposal, it dramatically improves the acceptance rate. This reduces the number of MH iterations needed to sufficiently mix the Markov chain.}
\begin{table*}[h]
	\centering
	\renewcommand{\arraystretch}{1.2}
	\scriptsize
	\setlength\aboverulesep{0.605mm}
	\setlength\belowrulesep{0.605mm}
	\begin{tabular}{ m{0.845\textwidth} >{\centering\arraybackslash\normalsize}m{0.10\textwidth} }
		\toprule
		\centering\textbf{Proposal} & \scriptsize\textbf{Probability of selecting proposal} \tabularnewline \toprule
			Select a grounded atomic axiom (e.g. $\color{RoyalBlue}\texttt{square}(c_1)$) and propose to replace it with an instantiation of a universal quantification (e.g. $\color{RoyalBlue}\forall x(\texttt{rectangle}(x) \land \texttt{rhombus}(x) \to \texttt{square}(x))$), where the antecedent conjuncts are selected uniformly at random from the other grounded atomic axioms for the constant $\color{RoyalBlue}c_1$: $\color{RoyalBlue}\texttt{rectangle}(c_1)$, $\color{RoyalBlue}\texttt{rhombus}(c_1)$, etc. &
			$\frac{1}{N}$ \tabularnewline \midrule
			The inverse of the above proposal: select an instantiation of a universal quantification and replace it with a grounded atomic axiom. &
			$\frac{1}{N}$ \tabularnewline \midrule
			Select an axiom that declares the size of a set (e.g. of the form $\color{RoyalBlue}\texttt{size}(\texttt{us\_states})=50$), and propose to change the size of the set by sampling from the prior distribution, conditioned on the maximum and minimum consistent set size. &
			$\frac{1}{N}$ \tabularnewline \midrule
			Select a node from a proof tree of type $\lor\text{I}$, $\to\hspace{-0.2em}\text{I}$, or $\exists\text{I}$.\footnote{\label{footnote:negated_conjunction}Also disproofs of conjunctions, if using classical logic.} These nodes were created in algorithm \ref{alg:init_proof} on lines \ref{alg:init_proof:disjunction_intro}, \ref{alg:init_proof:implication_intro}, and \ref{alg:init_proof:existential_intro}, respectively, where for each node, a single premise was selected out of a number of possible premises. This proposal naturally follows from the desire to explore other selections by re-sampling the proof: it simply calls \texttt{init\_proof} again on the formula at this proof node. &
			$\frac{1}{N}$ \tabularnewline \midrule
			\textbf{Merge:} Select a ``mergeable'' event; that is, three constants $(c_i,c_j,c_k)$ such that ${\color{RoyalBlue}\texttt{arg1}(c_i)=c_j}$, ${\color{RoyalBlue}\texttt{arg2}(c_i)=c_k}$, and $\color{RoyalBlue}t(c_i)$ for some constant $\color{RoyalBlue}t$ are axioms, and there also exist constants $(c_{i'},c_{j'},c_{k'})$ such that $i' > i$, ${\color{RoyalBlue}\texttt{arg1}(c_{i'})=c_{j'}}$, ${\color{RoyalBlue}\texttt{arg2}(c_{i'})=c_{k'}}$, and $\color{RoyalBlue}t(c_{i'})$ are axioms. Next, propose to merge $\color{RoyalBlue}c_{i'}$ with $\color{RoyalBlue}c_i$ by replacing all instances of $\color{RoyalBlue}c_{i'}$ with $\color{RoyalBlue}c_i$ in the proof trees, $\color{RoyalBlue}c_{j'}$ with $\color{RoyalBlue}c_j$, and $\color{RoyalBlue}c_{k'}$ with $\color{RoyalBlue}c_k$. This proposal is not necessary in that these changes are reachable with other proposals, but those proposals may have low probability, and so this can help to more easily escape local maxima. &
			$\frac{\alpha}{N}$ \tabularnewline \midrule
			\textbf{Split:} The inverse of the above proposal. &
			$\frac{\beta}{N}$ \tabularnewline \bottomrule
	\end{tabular}
	\vspace{-4mm}
	\caption{A list of the Metropolis-Hastings proposals implemented in \textsf{PWL} thus far. $N$, here, is a normalization term: $N = |A| + |U| + |C| + |P| + \alpha |M| + \beta |S|$ where: $A$ is the set of grounded atomic axioms in $T$ (e.g. $\color{RoyalBlue}\texttt{square}(c_1)$), $U$ is the set of universally-quantified axioms that can be eliminated by the second proposal, $C$ is the set of axioms that declare the size of a set (e.g. $\color{RoyalBlue}\texttt{size}(A)=4$), $P$ is the set of nodes of type $\lor\text{I}$, $\to\hspace{-0.2em}\text{I}$, or $\exists\text{I}$\cref{footnote:negated_conjunction} in the proofs $\bm{\pi}$, $M$ is the set of ``mergeable'' events (described above), and $S$ is the set of ``splittable'' events. In our experiments, $\alpha=2$ and $\beta=0.001$.}
	\vspace{-5mm}
	\label{table:mh_proposals}
\end{table*}
\end{savenotes}

\pagebreak\noindent ry by a sequence of mutations, then with sufficiently many iterations, the samples $\smash{T^{(t)}}$ and $\smash{\pi_i^{(t)}}$ will be distributed according to the true posterior $p(T,\bm{\pi}| \bm{x})$. If only a subset of possible theories and proofs are reachable from the initial theory, the MH samples will be distributed according to the true posterior \emph{conditioned} on that subset. This may suffice for many applications, particularly if the theories in the subset have desirable properties such as better tractability. But the subset cannot be too small since then \textsf{PWL} would lose generality.

The function \texttt{init\_proof} in algorithm \ref{alg:init_proof} recursively calls \texttt{init\_disproof}. Due to space limitations, we refer the reader to our code for this function; it closely mirrors the structure of \texttt{init\_proof}. The purpose of \texttt{init\_proof} is to find \emph{some} proof of a given higher-order formula, or return null if none exists. Its task is abduction, which is easier than theorem proving, since it can create new axioms as needed. The returned proof need not be ``optimal'' since it serves as the initial state for MH, which will further refine the proof. The validity of the proofs is guaranteed by the fact that \texttt{init\_proof} only returns valid proofs and the MH proposals preserve validity.

\vspace{-1mm}
\subsection{Language module}

For the language module, \textsf{PWM} uses the probabilistic model of \citet{DBLP:conf/conll/SaparovSM17}. The generative nature of their semantic parsing model allows it to fit seamlessly into \textsf{PWM} and \textsf{PWL}. The logical forms in their model are distributed according to a \emph{semantic prior}, which we replace with our distribution of logical forms conditioned on the theory $p(\pi_i | T)$. Their parser is probabilistic and finds the $k$-best logical forms that maximize $p(y_i|x_i,T)$ for a given input sentence. Combined with our reasoning module's ability to compute the probability of a logical form, the parser can resolve ambiguous interpretations of sentences by exploiting acquired knowledge. We will demonstrate the utility of this property in resolving lexical ambiguity.

However, the semantic grammar in \citet{DBLP:conf/conll/SaparovSM17} was designed for a \textsc{Datalog} representation of logical forms. Thus, we designed and implemented a new grammar for our more domain-general formalism in higher-order logic. Though their model induces preterminal production rules from data (e.g. N $\rightarrow$ ``cat''), we must manually specify the nonterminal production rules (e.g. NP $\rightarrow$ ADJP NP). This allows us to encode prior knowledge of the English language into \textsf{PWM}, dramatically improving its statistical efficiency and obviating the need for massive training sets to learn English syntax. It is nonetheless tedious to design these rules while maintaining domain-generality. Once specified, however, these rules can be re-used in new tasks and domains with minimal or no changes. We also improved their model to generalize over inflected forms of words. In the generative process, instead of generating sentence tokens directly (e.g. ``I am sleeping''), \textsf{PWM} generates word roots with flags indicating their inflection (e.g. ``I be[\textsc{1st},\textsc{sg}] sleep[\textsc{prs},\textsc{ptcp}]''). During parsing, this has the effect of performing morphological and semantic parsing jointly. We extracted the necessary comprehensive morphology information from Wiktionary \cite{Wiktionary}.

We train this new grammar to learn the parameters that govern the conditional distributions and the preterminal production rules. To do so, we construct a small \emph{seed training set} consisting of 55 labeled sentences, 47 nouns, 55 adjectives, and 20 verbs.\footnote{The grammar, morphology data, code, as well as the seed training set are available in our Github repository.} We wrote and labeled these sentences by hand, largely in the domain of astronomy, with the aim to cover a diverse range of English syntactic constructions. This small training set was sufficient thanks to the statistical efficiency of \textsf{PWM}. 

While \textsf{PWL} uses the same parsing algorithm of \citet{DBLP:conf/conll/SaparovSM17}, we provide an easier-to-understand presentation. Given an input sentence $y_i$, the parser aims to find the logical form(s) $x_i$ and derivation trees $t_i$ that maximize the posterior probability $p(x_i,t_i | y_i, T)$. This discrete optimization is performed using \emph{branch-and-bound} \cite{Land1960}: the algorithm starts by considering the set of all derivation trees and partitions it into a number of subsets (the ``branch'' step). For each subset $S$, the parser computes an upper bound on the log probability of any derivation in $S$ (the ``bound'' step). Having computed the bound for each subset, the parser puts them into a priority queue, prioritized by the bound. The parser then dequeues the subset with the highest bound and repeats this process, further subdividing this set, computing the bound for each subdivision, and adding them to the queue. Eventually, the parser will dequeue a subset containing a single derivation whose log probability is at least the highest priority in the queue. This derivation is optimal. The algorithm can be continued to obtain the top-$k$ derivations/logical forms. Since this algorithm operates over \emph{sets} of logical forms (where each set is possibly infinite), we implemented a data structure to sparsely represent such sets of higher-order formulas, as well as algorithms to perform set operations, such as intersection and subtraction.

\vspace{-0.4em}
\section{Experiments} \label{sec:experiments}

\begin{table*}[h]
	\vspace{-3mm}
	\setlength\extrarowheight{1.7mm}
	\setlength\aboverulesep{0.605mm}
	\setlength\belowrulesep{0.605mm}
	\renewcommand{\arraystretch}{0.65}
	\centering\scriptsize
 	\begin{tabular}{ l @{\hskip 6mm} c @{\hskip 3mm} | @{\hskip 3mm} c c >{\columncolor[rgb]{0.73,0.87,0.53}}c >{\columncolor[rgb]{0.73,0.87,0.53}}c} \toprule
		\textsc{Section}		& $N$ & \textsf{ProofWriter-All} & \textsf{ProofWriter-Iter} & \cellcolor{white}\textsf{PWL} (classical) & \cellcolor{white}(intuitionistic) \\ \midrule
		\textsf{Electricity1}	& 162	& 98.15				& 98.77 & 92.59				& \textbf{100.00} \\
		\textsf{Electricity2}	& 180	& 91.11				& 90.00 & 90.00				& \textbf{100.00} \\
		\textsf{Electricity3}	& 624	& 91.99				& 94.55 & 88.46				& \textbf{100.00} \\
		\textsf{Electricity4}	& 4224	& 91.64				& 99.91 & 94.22				& \textbf{100.00} \\
		\textsf{Birds1}			& 40	& \textbf{100.00}	& 95.00 & \textbf{100.00}	& \textbf{100.00} \\
		\textsf{Birds2}			& 40	& \textbf{100.00}	& 95.00 & \textbf{100.00}	& \textbf{100.00} \\ \midrule
		Average	& 5270	& 91.99				& 98.82 & 93.43				& \textbf{100.00} \\ \bottomrule
	\end{tabular} \\
	\vspace{-3mm}
	\caption{Zero-shot accuracy of \textsf{PWL} and baselines on the \textsf{ProofWriter} dataset.} \label{fig:ruletaker_results}
	\vspace{-6mm}
\end{table*}

\vspace{-0.2em}
\subsection{\textsf{ProofWriter}}
\vspace{-0.2em}

To demonstrate our implementation as a proof-of-concept, we evaluate it on two question-answering tasks. The first is the \textsf{ProofWriter} dataset \cite{DBLP:conf/acl/TafjordDC21}, which itself is based on the earlier \textsf{RuleTaker} dataset \cite{DBLP:conf/ijcai/ClarkTR20}. To evaluate and demonstrate the out-of-domain language understanding and reasoning ability of \textsf{PWL}, we use the \textsf{Birds-Electricity} ``open-world''\footnote{The dataset comes in two flavors: one which makes the closed-world assumption, and one that does not.} portion of the dataset, as the authors evaluated their method on this portion zero-shot, just as we do (i.e. the algorithm did not see any example from this portion during training). This portion of the data is subdivided into 6 sections, each with varying degrees of difficulty. An example from this dataset is shown in figure \ref{fig:proofwriter_example}. For each example, \textsf{PWL} reads the context and abduces a theory. Next, it parses the query sentence $y_{n+1}$ into a logical form $x_{n+1}$ and estimates its \emph{unnormalized} probability:
\begin{align}
	p(&x_{n+1}\mid \bm{x}) \propto p(x_1,\ldots,x_{n+1}), \\
		&= \sum_{T,\pi_i\text{ proof of }x_i} p(T) \prod_{i=1}^{n+1} p(\pi_i\mid T), \label{eq:exact_posterior_predictive} \\
		&\approx \hspace{-1.5em}\sum_{T^{(t)},\pi_i^{(t)}\sim T,\bm{\pi} \mid \bm{x}}\hspace{-1.5em} p(T^{(t)}) \prod_{i=1}^{n+1} p(\pi_i^{(t)}\mid T^{(t)}).
\end{align}
Here, $\bm{x}$ are the previously read logical forms (the context). Since the quantity in equation \ref{eq:exact_posterior_predictive} is intractable to compute, \textsf{PWL} approximates it by sampling from the posterior $T,\pi_1,\ldots,\pi_{n+1} \mid x_1,\ldots,x_{n+1}$ and summing over distinct samples. Although this approximation seems crude, the sum is dominated by a small number of the most probable theories and proofs, and MH is an effective way to find them, as we observe in experiments. MH is run for $400$ iterations, and at every $100^{\text{\scriptsize th}}$ iteration, \textsf{PWL} re-initializes the Markov chain by performing $20$ ``exploratory'' MH steps (i.e. consisting of only the third and fourth proposals in table \ref{table:mh_proposals} and accepting every proposal). This re-initialization is analogous to a random restart and can help to escape from local maxima. However, it may be promising to explore other approaches to compute this quantity, such as \citet{DBLP:conf/iclr/LuoB00DAC20}. Once \textsf{PWL} has computed this probability for the query sentence, it does the same for the negation of the sentence. These unnormalized probabilities are compared, and if they are within $2000$ in log probability, \textsf{PWL} returns the label \textsf{unknown}. If the first probability is sufficiently larger than the second, \textsf{PWL} returns \textsf{true}, and otherwise, return \textsf{false}. The parameters in the prior were set by hand initially by choosing values which we thought were reasonable (e.g. the average length of a natural deduction proof for a sentence containing a simple subject noun phrase, object noun phrase, and transitive verb is around 20 steps, which is why the Poisson parameter for the proof length is set to 20). The values were tweaked as necessary by running the algorithm on toy examples during debugging. Note that the sentences ``Bill is a bird'' and ``Bill is not a bird'' can still both be true if each ``Bill'' refers to distinct entities. To avoid this, we chose an extreme value of the prior parameter such that the log prior probability of a theory with two entities having the same name is $2000$ less than that of a theory where the name is unique. It is for this reason $2000$ was chosen as the threshold for determining whether a query is \textsf{true}/\textsf{false} vs \textsf{unknown}. This prior worked well enough in our experiments, but the goal is to have a single prior work well with any task, so further work to explore which priors work better across a wider variety of tasks is welcome. We evaluated \textsf{PWL} using both classical and intuitionistic logic, even though the ground truth labels in the dataset were generated using \emph{intuitionistic logic}.

\begin{figure}
\scriptsize
	\vspace{1mm}
	\renewcommand{\arraystretch}{0.9}
	\begin{tabular}{m{0.45\textwidth}} \toprule
	\textbf{Context:} ``The switch is on. The circuit has the bell. If the circuit has the switch and the switch is on then the circuit is complete. If the circuit does not have the switch then the circuit is complete. If the circuit is complete and the circuit has the light bulb then the light bulb is glowing. If the circuit is complete and the circuit has the bell then the bell is ringing. If the circuit is complete and the circuit has the radio then the radio is playing.'' \\ \midrule
	\textbf{Query:} ``The circuit is complete.'' \hspace{1em} \textbf{\textsf{true}}, \textbf{\textsf{false}}, \textbf{\textsf{unknown}?} \\ \bottomrule
	\end{tabular}
	\vspace{-2mm}
	\caption{An example from the \textsf{Electricity1} section in the \textsf{ProofWriter} dataset. Its label is \textsf{unknown}. Under classical logic, the query is provably true from the information in the 1st, 3rd, and 4th sentences.} \label{fig:proofwriter_example}
	\vspace{-2em}
\end{figure}

Table \ref{fig:ruletaker_results} lists the zero-shot accuracy of \textsf{PWL}, comparing with baselines based on the T5 transformer \cite{DBLP:journals/jmlr/RaffelSRLNMZLL20}. We emphasize here that \textsf{PWL} is not perfectly comparable to the baseline, since they aim to demonstrate that their method can \emph{learn} to reason. We instead aim to demonstrate that \textsf{PWL}'s ability to parse and reason end-to-end generalizes to an out-of-domain question-answering task. The baseline is trained on other portions of the \textsf{ProofWriter} data, whereas \textsf{PWL} is trained only on its seed training set. \textsf{PWL} performed much better using intuitionistic logic than classical logic, as expected since the ground truth labels were generated using intuitionistic semantics. However, most real-world reasoning tasks would take the law of the excluded middle to be true, and classical logic would serve as a better default. Although the task is relatively simple, it nevertheless demonstrates the proof-of-concept and the promise of further research.

\newcommand\mytab[1]{\begin{tabular}[t]{l}#1\end{tabular}}
\newcommand\rotbox[2]{\setlength\extrarowheight{0mm}\rotatebox{#1}{\mytab{#2}\hspace{-3mm}}\hspace{-2mm}\setlength\extrarowheight{2mm}}

\begin{table*}[h]
	\vspace{-3mm}
	\scriptsize
	\setlength\extrarowheight{2mm}
	\setlength\aboverulesep{0.605mm}
	\setlength\belowrulesep{0.605mm}
	\renewcommand{\arraystretch}{0.62}
	\begin{center}
 	\begin{tabular}{ p{9.5em} | >{\raggedleft\arraybackslash}p{1.8em} | >{\raggedleft\arraybackslash}p{1.8em} >{\raggedleft\arraybackslash}p{1.8em} >{\raggedleft\arraybackslash}p{1.8em} >{\raggedleft\arraybackslash}p{1.8em} >{\raggedleft\arraybackslash}p{1.8em} >{\raggedleft\arraybackslash}p{1.65em} >{\raggedleft\arraybackslash}p{1.65em} >{\raggedleft\arraybackslash}p{1.65em} >{\raggedleft\arraybackslash}p{1.65em} >{\raggedleft\arraybackslash}p{1.65em} >{\raggedleft\arraybackslash}p{1.65em} >{\raggedleft\arraybackslash}p{1.65em} >{\raggedleft\arraybackslash}p{1.65em}} \toprule
								& \rotbox{65}{\textbf{all}} & \rotbox{65}{superlative} & \rotbox{65}{subjective \\[0mm] concept def.} & \rotbox{65}{objective \\[0mm] concept def.} & \rotbox{65}{lexical \\[0mm] ambiguity} & \rotbox{65}{negation} & \rotbox{65}{large \\[0mm] context} & \rotbox{65}{arithmetic} & \rotbox{65}{counting} & \rotbox{65}{0 subsets} & \rotbox{65}{1 subset} & \rotbox{65}{2 subsets} & \rotbox{65}{3 subsets} & \rotbox{65}{4 subsets} \\[-2mm] \midrule
		$N$										& \centering 600	& \centering 210	& \centering 150	& \centering 170	& \centering 180	& \centering 102	& \centering 100	& \centering 20	& \centering 30	& \centering 85	& \centering 213	& \centering 187	& \centering 85	& 30\phantom{-} \\ \midrule
		\textsf{UnifiedQA}						& 33.8	& 29.5	& 7.3	& 33.5	& 32.8	& 14.7	& 43.0	& 10.0	& \textbf{20.0}	& 41.2	& 47.9	& 27.8	& 8.2	& \textbf{23.3} \\
		\textsf{Boxer} + \textsf{E}				& 9.7	& 0.0	& 12.0	& 11.8	& 0.0	& 15.7	& 14.0	& 10.0	& 0.0	& 7.1	& 17.8	& 5.3	& 4.7	& 0.0 \\
		\textsf{Boxer} + \textsf{Vampire}		& 9.7	& 0.0	& 12.0	& 11.8	& 0.0	& 15.7	& 14.0	& 10.0	& 0.0	& 7.1	& 17.8	& 5.3	& 4.7	& 0.0 \\
		\textsf{PWL} parser + \textsf{E}		& 5.0	& 0.0	& 13.3	& 2.9	& 0.0	& 15.7	& 4.0	& 10.0	& 0.0	& 1.2	& 7.0	& 5.3	& 4.7	& 0.0 \\
		\textsf{PWL} parser + \textsf{Vampire}	& 9.0	& 0.0	& 13.3	& 11.2	& 0.0	& 15.7	& 4.0	& 10.0	& 0.0	& 12.9	& 13.6	& 5.3	& 4.7	& 0.0 \\
		\rowcolor[rgb]{0.73,0.87,0.53}\textsf{PWL}	& \textbf{43.1}	& \textbf{40.5}	& \textbf{33.3}	& 33.5	& \textbf{34.4}	& \textbf{23.5}	& \textbf{45.0}	& 10.0	& 0.0	& \textbf{43.5}	& \textbf{62.9}	& \textbf{39.0}	& \textbf{17.6}	& 0.0 \\ \bottomrule
	\end{tabular} \\
	\end{center}
	\vspace{-1mm}
	\textsc{Legend:} \textbf{superlative} The subset of the dataset with examples that require reasoning over superlatives, i.e. ``longest river.'' \\
	\textbf{subjective concept def.} Subset with definitions of ``subjective'' concepts, i.e. ``Every river longer than 500 km is {\color{mdtRed}major}.'' \\
	\textbf{objective concept def.} Subset with definitions of ``objective'' concepts, i.e. the {\color{mdtRed}population} of a location is the number of people living there. \\
	\textbf{lexical ambiguity} Subset with lexical ambiguity, i.e. ``has'' means different things in ``a state has a city named'' vs ``a state has an area of...'' \\
	\textbf{negation} Subset with examples that require reasoning with classical negation (negation-as-failure is insufficient). \\
	\textbf{large context} Subset of examples where there are at least 100 sentences in the context. \\
	\textbf{arithmetic} Subset with examples that require simple arithmetic. \textbf{counting} Subset with examples that require counting. \\
	\textbf{$\bm{n}$ subset(s)} Examples that belong to exactly $n$ of the above subsets (no example is a member of more than 4 subsets).
	\vspace{-4mm}
	\caption{Zero-shot accuracy of \textsf{PWL} and baselines on the \textsf{FictionalGeoQA} dataset.} \label{fig:fictionalgeoqa_results}
	\vspace{-5mm}
\end{table*}

\subsection{\textsf{FictionalGeoQA}}
\vspace{-0.3em}

The sentences in the \textsf{ProofWriter} experiment are template-generated and have simple semantics. For the sake of evaluation more representative of real-world language, we introduce a new question-answering dataset called \textsf{FictionalGeoQA}\footnote{Available at \href{https://github.com/asaparov/fictionalgeoqa}{\texttt{\footnotesize github.com/asaparov/fictionalgeoqa}}}. To create this dataset, we took questions from \textsf{GeoQuery} \cite{DBLP:conf/aaai/ZelleM96}, and for each question, we wrote a paragraph context containing the information necessary to answer the question. We added distractor sentences to make the task more robust against heuristics. Whenever possible, the sentences in this paragraph were taken from Simple English Wikipedia. However, some facts, such as the lengths of rivers, are not expressed in sentences in Wikipedia (they typically appear in a table on the right side of the page), so we wrote those sentences by hand: We took questions from \textsc{GeoQuery} that expressed the desired fact in interrogative form (e.g. ``What is the length of \texttt{<river name>}?'') and converted them into declarative form (e.g. ``The length of \texttt{<river name>} is \texttt{<length>}.''). The resulting dataset contains 600 examples, where 67.4\% of the sentences are from Simple English Wikipedia, and 90\% of the examples contain at least one sentence \emph{not} from Wikipedia. We replaced all place names with fictional ones to remove any confounding effects from pretraining. To keep the focus of the evaluation on reasoning ability, we chose to restrict the complexity of the language. In particular, each sentence is independent and can be understood in isolation (e.g. no cross-sentential anaphora). The sentences \emph{are} more complex than those in \textsf{ProofWriter}, having more of the complexities of real language, such as synonymy, lexical ambiguity (e.g. what is the semantics of ``has'' in ``a state has city'' vs ``a state has area''; or whether ``largest state'' refers to area or population), and syntactic ambiguity. This increased difficulty is evident in the results. This dataset is meant to evaluate out-of-domain generalizability, so we do not provide a separate training set for fine-tuning.

\begin{figure}
\scriptsize
	\renewcommand{\arraystretch}{0.9}
	\begin{tabular}{m{0.45\textwidth}} \toprule
	\textbf{Context:} ``River Giffeleney is a river in Wulstershire. River Wulstershire is a river in the state of Wulstershire. River Elsuir is a river in Wulstershire. The length of River Giffeleney is 413 kilometers. The length of River Wulstershire is 830 kilometers. The length of River Elsuir is 207 kilometers. Every river that is shorter than 400 kilometers is not major.'' \\ \midrule
	\textbf{Query:} ``What rivers in Wulstershire are not major?'' \\ \bottomrule
	\end{tabular}
	\vspace{-2mm}
	\caption{An example from \textsf{FictionalGeoQA}, a new fictional geography question-answering dataset that we created to evaluate reasoning in natural language understanding.} \label{fig:georeasoning_example}
	\vspace{-5mm}
\end{figure}

We compare \textsf{PWL} (using classical logic) with a number of baselines: (1) \textsf{UnifiedQA} \cite{DBLP:conf/emnlp/KhashabiMKSTCH20}, a QA system based on large-scale neural language models, (2) \textsf{Boxer} \cite{DBLP:conf/nodalida/Bos15}, a wide-coverage semantic parser, combined with \textsf{Vampire} 4.5.1 \cite{DBLP:conf/cav/KovacsV13}, a theorem prover for full first-order logic, (3) \textsf{Boxer} combined with \textsf{E} 2.6 \cite{DBLP:conf/cade/0001CV19}, another theorem prover for full first-order logic, (4) the language module of \textsf{PWL} combined with \textsf{Vampire}, and (5) the language module of \textsf{PWL} combined with \textsf{E}. The results are shown in figure \ref{fig:fictionalgeoqa_results}, along with a breakdown across multiple subsets of the dataset. \textsf{UnifiedQA} performs relatively well but fairs more poorly on questions with negation and subjective concept definitions (e.g. ``Every river longer than 500km is major... What are the major rivers?''). Humans are easily able to understand and utilize such definitions, and the ability to do so is instrumental in learning about new concepts or words in new domains. \textsf{PWL} is able to fare better than \textsf{UnifiedQA} in examples with lexical ambiguity, as a result of the language module's ability to exploit acquired knowledge to resolve ambiguities. We find that \textsf{Boxer} has significantly higher coverage than \textsf{PWL} ($100\%$ vs $79.8\%$) but much lower precision. For instance, \textsf{Boxer} uses the semantic representation in the Parallel Meaning Bank \cite{DBLP:conf/eacl/BosEBAHNLN17} which has a simpler representation of superlatives, and is thus unable to capture the correct semantics of superlatives in examples of this dataset. We also find that for most examples, \textsf{Boxer} produces different semantics for the question vs the context sentences, oftentimes predicting the incorrect semantic role for the interrogative words, which leads to the theorem provers being unable to find a proof for these extra semantic roles. We also experimented with replacing our reasoning module with a theorem prover and found that for almost all examples, the search of the theorem prover would explode combinatorially. This was due to the fact that our semantic representation relies heavily on \emph{sets}, so a number of simple set theoretic axioms are required for the theorem provers, but this quickly causes the deduction problem to become undecideable. Our reasoning module instead performs abduction, and is able to create axioms to more quickly find an initial proof, and then refine that proof using MH. Despite our attempt to maximize the generalizability of the grammar in \textsf{PWL}, there are a number of linguistic phenomena that we did not yet implement, such as interrogative subordinate clauses, wh-movement, spelling or grammatical mistakes, etc, and this led to the lower coverage on this dataset. Work remains to be done to implement these missing production rules in order to further increase the coverage of the parser.

\section{Conclusions and future work}

We introduced \textsf{PWM}, a fully-symbolic Bayesian model of semantic parsing and reasoning, which we hope serves as a compelling first step in a research program toward more domain- and task-general NLU. We derived \textsf{PWL}, an efficient inference algorithm that reads sentences by parsing and abducing updates to its latent world model that capture the semantics of those sentences, and empirically demonstrated its ability to generalize to two out-of-domain question-answering tasks. To do so, we created a new question-answering dataset, \textsf{FictionalGeoQA}, designed specifically to evaluate reasoning ability while capturing more of the complexities of real language and being robust against heuristic strategies. \textsf{PWL} is able to read and understand sentences with richer semantics, such as definitions of new concepts. In contrast with past deductive reasoning approaches, \textsf{PWL} performs abduction, which is computationally easier. The highly underspecified nature of the problem of abduction is alleviated by the probabilistic nature of \textsf{PWL}, as it gives a principled way to find the most probable theories. We present an inference strategy where Metropolis-Hastings (MH) is performed on each sentence, in sequence, where the previous sample of the theory and proofs provides a warm-start for inference of the next sentence, reducing the number of MH iterations.

There are many avenues for future work: A simple prior was used for proofs $p(\pi_i| T)$, and an alternate approach could be to use a compositional exchangeable distribution such as adaptor grammars \cite{DBLP:conf/nips/JohnsonGG06}.

The first MH proposal in table \ref{table:mh_proposals} is simple but restrictive: the antecedent conjuncts and the consequent are restricted to be atomic. MH would be able to explore a much larger and semantically richer set of theories if the antecedent or consequent could contain more complex formulas, including quantified formulas. In addition, the inference algorithm sometimes becomes stuck in local maxima. One way to improve the efficiency of inference is to add a new MH proposal that specifically proposes to split or merge types. For example, if the theory has the axioms $\color{RoyalBlue}\texttt{cat}(c_1)$ and $\color{RoyalBlue}\texttt{dog}(c_1)$, this proposal would split $c_1$ into two concepts: $\color{RoyalBlue}\texttt{cat}(c_1)$ and $\color{RoyalBlue}\texttt{dog}(c_2)$. This kind of type-based Markov chain Monte Carlo is similar in principle to \citet{DBLP:conf/naacl/LiangJK10}.

As mentioned earlier, a model of context is necessary in the language module to properly handle cross-sentential anaphora and conversational contexts. Real language very rarely consists of sentences that are independent of context. There are also many research questions on the issue of \emph{scalability}. Although \textsf{PWL} is able to scale to examples in \textsf{FictionalGeoQA} with more than 100 sentences, there are two main bottlenecks currently preventing it from scaling to significantly larger theories: (1) the maintenance of global consistency, and (2) the unfocused nature of the current MH proposals. When checking for consistency of a new axiom, rather than considering all other axioms/sets in the theory, it would be preferable to only consider the portion of the theory relevant to the new axiom. Additionally, the current MH proposals do not take into account the goal of reasoning. For example, if the current task is to answer a question about geography, then MH proposals for proofs unrelated to geography are wasteful, and would increase the number of MH steps needed. A more clever goal-aware approach for selecting proofs to mutate would help to alleviate this problem and improve scalability. \textsf{PWM} provides a path to incorporate information from additional modalities in principled fashion: for example by adding a generative model of images, which would serve as a separate ``vision module.'' In addition, even though \textsf{PWL} is fully-symbolic, non-symbolic methods could be used for expressive prior/proposal distributions or approximate inference. There are many fascinating research paths to pursue from here.

\section*{Acknowledgments}

We thank the anonymous reviewers and the Action Editor for their invaluable feedback. We also thank Peter Clark, William W. Cohen, Rik van Noord, and Johan Bos for their insightful and helpful discussion.

\bibliography{tacl2018v2}
\bibliographystyle{acl_natbib}

\end{document}